\documentclass[]{spie} 
\usepackage{times}  % DO NOT CHANGE THIS
\usepackage{helvet}  % DO NOT CHANGE THIS
\usepackage{courier}  % DO NOT CHANGE THIS
\usepackage[hyphens]{url}  % DO NOT CHANGE THIS
\usepackage{graphicx} % DO NOT CHANGE THIS
\usepackage{caption} % DO NOT CHANGE THIS AND DO NOT ADD ANY OPTIONS TO IT
\usepackage{amsmath}
\usepackage{amsfonts}       % blackboard math symbols
\usepackage{algorithm}
\usepackage{algorithmic}
\usepackage{newfloat}
\usepackage{listings}
\DeclareCaptionStyle{ruled}{labelfont=normalfont,labelsep=colon,strut=off} % DO NOT CHANGE THIS
\floatstyle{ruled}
\newfloat{listing}{tb}{lst}{}
\floatname{listing}{Listing}
\newcommand{\comment}[1]{}

\RequirePackage{color, soul}
\RequirePackage{xcolor}
\RequirePackage[normalem]{ulem} % for \sout{}

\title{Extracting Explanations, Justification, and Uncertainty \\ from Black-Box Deep Neural Networks}
\author[a]{
    Paul Ardis
    }
\author[b]{
    Arjuna Flenner
    }
\affil[a]{
    GE Aerospace Research,
    1 Research Circle,
    Niskayuna, NY 12309, USA
    }
\affil[b]{
    GE Aerospace,
    3290 Patterson Avenue SE,
    Grand Rapids, MI 49512, USA
    }
\begin{document}

\maketitle

\begin{abstract}
Deep Neural Networks (DNNs) do not inherently compute or exhibit empirically-justified task confidence. In mission critical applications, it is important to both understand associated DNN reasoning and its supporting evidence. In this paper, we propose a novel Bayesian approach to extract explanations, justifications, and uncertainty estimates from DNNs. Our approach is efficient both in terms of memory and computation, and can be applied to any black box DNN without any retraining, including applications to anomaly detection and out-of-distribution detection tasks. We validate our approach on the CIFAR-10 dataset, and show that it can significantly improve the interpretability and reliability of DNNs.

%    Explanation: Why did the model make this prediction?
%    Justification: Why should I trust this prediction?
%    Uncertainty: How confident is the model in this prediction?
\end{abstract}
\section{Introduction}

\begin{figure}[t!]
    \centering
    \includegraphics[width=\linewidth]{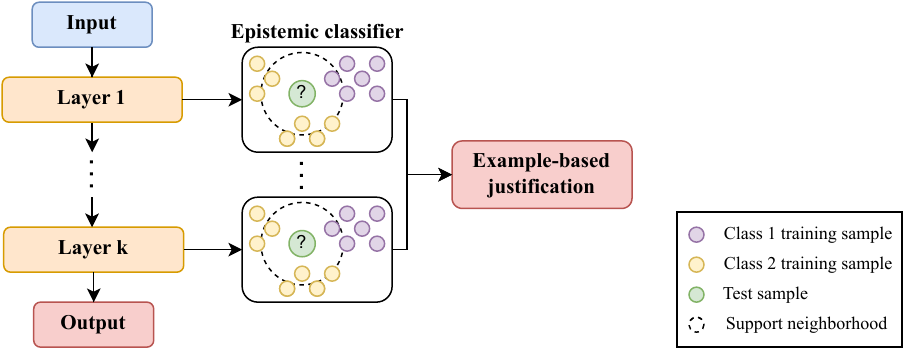}
    \vspace{0.2cm}
    \caption{An illustration of the example-based XAI approach of Virani et al.~\cite{ViraniIY20}. Their approach uses the transformation spaces from intermittent layers of a pre-trained DNN to evaluate the model’s justification and builds a support neighborhood around each test sample's transformed point.}
    \label{fig:epi_classifiers}
\end{figure}

Deep neural networks (DNNs) are powerful machine learning models that can learn complex patterns from data. However, their complexity can make it difficult to understand why a DNN makes a particular prediction purely from their mathematical construction. We build upon the prior Explainable AI (XAI) work from Virani et al.~\cite{ViraniIY20} (see Figure \ref{fig:epi_classifiers}) which effectively explains if there is proper training data to support the decisions. However, their memory and computational footprint prevents the approach from being effective for very large data sets and on edge devices. We exploit the concept of Sparse Gaussian processes~\cite{SnelsonG06, BuiYT17, Titsias09, HoangHL15} to overcome these two computational challenges while maintaining the their method's accuracy and explainability. Our approach is a computationally efficient XAI method that extracts example-based justifications and uncertainty estimates that can be applied to any pre-trained DNN.

Explainable artificial intelligence addresses the black-box problem in many ways: it decomposes the model construction for more intuitive human understanding; it supplements  models with tools that explain their decision-making; or a structured combination of the previous two. By providing insights into how AI systems make decisions, XAI can help humans to identify and correct potential errors before they cause harm. As such, XAI is a critical tool for ensuring the safety and human trust in mission-critical tasks such as jet engine maintenance and airport safety. In the context of jet engine maintenance, XAI can be used to provide detailed reasoning about the evidence suggesting potential problems before they can cause failures, recommend corrective actions to prevent potential issues, and monitor the performance of engines to ensure that they continue to operate safely. In the context of airport safety, XAI can be used to detect and explain potential hazards on the airport grounds, expedite human understanding of the evidence behind security threats, and monitor the movement of aircraft to ensure that they follow safe takeoff and landing directions.

While there are a number of competing standards and meta-studies of XAI~\cite{SchwalbeF23}, primary seminal papers explore XAI using one of three main approaches. The first approach is to use gradient-based methods to identify the input features that are most important for a particular prediction~\cite{AnconaCOG19}. This can be done by calculating the gradient of the loss function with respect to the input features - the features with the largest gradients are the ones that have the most influence on the prediction. The second approach to explainability is to train the DNN to output text that explains its reasoning. This can be done by using a technique called attention, which allows the DNN to focus on specific input features when making a prediction~\cite{Neely2021}. The DNN can then be trained to output text that describes the features that it focused on. The third approach to explainability is to train the DNN using metric learning~\cite{suarez2021}. This involves training the DNN to learn a distance metric that can be used to measure the similarity between input features. The metric can then be used to explicitly identify training set samples that are similar to any given test sample. The DNN can also then be used to make predictions for the testing sample based solely on the predictions that it made for the similar training samples.

Each of these approaches has its own advantages and disadvantages. Gradient-based methods are relatively simple to implement, but they can be difficult to interpret in their raw form. Text-based explanations can be more informative, but they can be more difficult to generate without adding a complex model of relevant human language. Metric learning can be used to generate both simple and informative explanations, but it can be more computationally expensive as it requires retraining the neural network.

We address the limitations of existing methods by presenting an approach to extract example-based justifications and uncertainty estimates from pre-trained DNNs. Our example-based justifications can be used to determine the training set samples that were the most relevant to any given test sample. In scenarios where there are an insufficient number of relevant training set samples, we can conclude that the neural network is extrapolating and has a high potential of misclassification. In addition, our uncertainty estimates provide explicit information about the model's prediction confidence based upon local exemplar density and coherence, and this uncertainty can be thresholded to limit operation to avoid potential misclassifications. Our contributions are as follows:
\begin{itemize}
    \item We present an approach that uses sparse Gaussian processes to take the latent embeddings from a pre-trained DNN and predict the model output. Such an approach will allow us to estimate the prediction uncertainty while also leveraging the DNN’s representation power.
    \item We obtain example-based justifications for our model predictions by selecting the training set samples that are highly correlated with the test samples. We use the SGP’s kernel functions to determine how the samples are correlated.
\end{itemize}

\subsection{Advantages of our approach}

\begin{itemize}
    \item Bayesian uncertainty quantification using sparse Gaussian process (SGPs~\cite{CandelaRW07, BuiYT17})
        \begin{itemize}
            \item SGPs have $\mathcal{O}(nm^2+m^3)$ computation cost~\cite{Titsias09}, where n is the number of training samples and m is the number of SGP inducing points
            \item SGPs can be trained on large datasets using stochastic gradient descent~\cite{WilkDJAAH20}
            \item SGPs support non-Gaussian likelihoods (e.g., softmax likelihood)
            \item SGPs can be defined on Riemannian manifolds and leverage geodesics to determine correlations~\cite{BorovitskiyTMD20}
        \end{itemize}
    \item Prediction justification from example-based explanations
        \begin{itemize}
            \item Reduce neighborhood operators' computation cost by leveraging SGP inducing points
        \end{itemize}
\end{itemize}
\section{Background: Gaussian Processes}

\textbf{Gaussian processes (GP)},~\cite{RasmussenW05} can be used as priors in a non-parametric Bayesian approach for regression and classification problems. Let $\mathbf{x} \in \mathbb{R}^d$, then a Gaussian process is uniquely defined using a mean function $m: \mathbb{R}^d \rightarrow \mathbb{R}$ and covariance function $k:\mathbb{R}^d \times \mathbb{R}^d \rightarrow \mathbb{R}$. The covariance function $k$ must be a positive-definite kernel function which may depend on a set of parameters $\Theta$. Given these two functions, a stochastic process $f \in \mathbb{R}$ indexed by $\mathbf{x} \in \mathbb{R}^d$ such that 
\begin{eqnarray*}
m(\mathbf{x}) &=& \mathbb{E}[f(\mathbf{x})], \\
k(\mathbf{x}_1, \mathbf{x}_2) &=& \mathbb{E}[(f(\mathbf{x}_1) - m(\mathbf{x}_1))(f(\mathbf{x}_2) - m(\mathbf{x}_2))]
\end{eqnarray*}
is a GP. If $f$ is a Gaussian process with mean $m(\cdot)$ and covariance kernel $k(\cdot, \cdot)$ write 
\begin{eqnarray*}
f(\mathbf{x}) &\sim& \mathcal{GP}(f | m(\mathbf{x}), k(\mathbf{x}, \mathbf{x})).
\end{eqnarray*}
For regression problems, the GP is often used to define a generative model of the form:
\begin{eqnarray*}
y_i &=& f(\mathbf{x}_i) + \epsilon_i, \\
\epsilon_i &\sim& \mathcal{N}(\epsilon | 0, \sigma_{\text{noise}}^2), \\
f(\mathbf{x}_i) &\sim& \mathcal{GP}(f | m(\mathbf{x}, k(\mathbf{x}, \mathbf{x})). 
\end{eqnarray*}
 
An essential concept in the following discussion is the number of variables $\mathbf{x}$ used in the calculations. The following notation is used to highlight the set sizes with the goal of illuminating the computational complexity. Let $\mathbf{X}_n = [\mathbf{x}_1, \mathbf{x}_2,...,\mathbf{x}_n]^\top$ be a set of locations. Define $1 \leq i,j \leq n$ and let $\mathbf{K}_{nn}$ be a matrix with elements $k(\mathbf{x}_i, \mathbf{x}_j)$. Note that the subscript $nn$ refers to the number of rows and number of columns and not a row or column index. Define the random vector $\mathbf{f}_n = [f(\mathbf{x}_1), f(\mathbf{x}_2),...,f(\mathbf{x}_n)]^\top$. Given the location $\mathbf{x}$, define $\mathbf{k}_{\mathbf{x}n} = [k(\mathbf{x}, \mathbf{x}_1), \dots, k(\mathbf{x}, \mathbf{x}_n)]^{\top}$. 

Bayesian methods assume a prior data model and calculate a posterior distribution once data is observed. Consider an observed data set $\mathcal{D}_n = \{(\mathbf{x}_i, y_i)\; \vert \; 1 \leq i \leq n, \; \mathbf{x}_i \in \mathbb{R}^n,\; y_i \in \mathbb{R}\}$. Define the vector $\mathbf{y} = [y_1, \dots, y_n]^{\top}$ and assume that the mean function $m(\cdot) = 0$. With this mean function, recall that the definition of a GP designates $p(\mathbf{f} | \mathbf{X}_n) = \mathcal{N}(\mathbf{f} | \mathbf{0}, \mathbf{K}_{nn})$.  Since all the random variables in the GP regression model are Gaussian distributed, the posterior distribution is also a GP with mean function $m_\mathbf{y}(\cdot)$ and covariance kernel $k_\mathbf{y}(\cdot, \cdot)$ given by 
\begin{equation}
\begin{aligned}
\label{eq:gp}
m_\mathbf{y}(\mathbf{x}) &= \mathbf{k}_{\mathbf{x}n}^{\top}(\mathbf{K}_{nn} + \sigma_{\text{noise}}^{2}I)^{-1}\mathbf{y}\,, \\
k_\mathbf{y}(\mathbf{x}, \mathbf{x}^\prime) &= k(\mathbf{x}, \mathbf{x}^\prime) - \mathbf{k}_{\mathbf{x}n}^{\top}(\mathbf{K}_{nn} + \sigma_{\text{noise}}^{2}I)^{-1}\mathbf{k}_{\mathbf{x}^\prime n}.
\end{aligned}
\end{equation}
The parameters $\{\Theta, \sigma_{\text{noise}}\}$ can be estimated by maximizing the log marginal likelihood 
\begin{eqnarray*}
\log p(\mathbf{y}) = \log [ \mathcal{N}(\mathbf{y} | \mathbf{0}, \mathbf{K}_{nn} + \sigma^2_{\text{noise}} I)].
\end{eqnarray*}

Inspection of the posterior parameters illustrates a computational bottleneck. Specifically, computation is via an inversion of a matrix of size $n \times n$, where $n$ is the cardinality of the data set, which is a $\mathcal{O}(n^3)$ operation. For large data sets, this is computationally infeasible. This observation has motivated the development of Sparse Gaussian processes. 

\textbf{Sparse Gaussian processes (SGPs)}~\cite{SnelsonG06, BuiYT17, Titsias09, HoangHL15} are designed to overcome the cubic nature of calculating GP posteriors by approximating the GP using another Gaussian process supported with $m \ll n$ \textit{inducing points}. This reduces the posterior calculation to a matrix inversion using a $m \times m$ matrix; SGPs reduce the computations to $\mathcal{O}(m^3)$.

This paper uses the sparse variational GP (SVGP)~\cite{Titsias09, BurtRW19, BauerWR16}. This approach implements a variational Bayesian method that approximates the true posterior distribution with a variational distribution $q$ defined using inducing points $\mathbf{X}_m$, mean parameter $\boldsymbol{\mu}$, and covariance parameter $\mathbf{A}$ as variational parameters. 

Once these parameters are calculated, they can be used to define an approximate GP with mean parameter $m_{\mathbf{y}}^q(\cdot)$ and covaraince function $k_{\mathbf{y}}^q(\cdot, \cdot)$ given by
\begin{equation}
\begin{aligned}
\label{eq:sgp}
    m_\mathbf{y}^q(\mathbf{x}) &= \mathbf{k}_{\mathbf{x}m}^\top \mathbf{K}_{mm}^{-1} \boldsymbol{\mu}\,, \\
    k_\mathbf{y}^q(\mathbf{x}, \mathbf{x}^\prime) &= k(\mathbf{x}, \mathbf{x}^\prime) - \mathbf{k}_{\mathbf{x}m}^{\top} \mathbf{K}_{mm}^{-1} \mathbf{k}_{m\mathbf{x}^\prime} \\
    & \quad + \mathbf{k}_{\mathbf{x}m}^\top \mathbf{K}_{mm}^{-1} \mathbf{A} \mathbf{K}_{mm}^{-1} \mathbf{k}_{m\mathbf{x}^\prime}\,.
\end{aligned}
\end{equation}
Given the inducing points $\mathbf{X}_m$, the parameters $\boldsymbol{\mu}$ and $\mathbf{A}$ can be calcualted as 
\begin{eqnarray*}
\boldsymbol{\mu} &=& \sigma_{\text{noise}}^{-2}\mathbf{K}_{mm} \boldsymbol{\Sigma} \mathbf{K}_{mn} \mathbf{y}, \\
\mathbf{A} &=& \mathbf{K}_{mm} \boldsymbol{\Sigma}\mathbf{K}_{mm}, \\
\boldsymbol{\Sigma} &=& (\mathbf{K}_{mm} + \sigma_{\text{noise}}^{-2} \mathbf{K}_{mn}\mathbf{K}_{nm})^{-1}. 
\end{eqnarray*}
There are many techniques to determine the inducing points $\mathbf{X}_m$; See Titsias et al.\cite{Titsias09} for more details. 

We point out that the subscripts of $\mathbf{K}_{mn}$ highlights if the inducing points or the data correspond to the first and second component of the covariance function. For example, $\mathbf{K}_{mn}$ varies the inducing point across the rows and varies the data points across the columns.  

Bauer et al.~\cite{BauerWR16} provides an in-depth analysis of the SVGP's lower bound which illuminates role of the inducing points. The variational approximation maximizes the  evidence lower bound~(ELBO)~$\mathcal{F}$ with regard to the variational distribution:
\begin{equation}
\label{vfe}
\begin{aligned}
    \mathcal{F} &= \underbrace{\frac{n}{2} \log(2\pi)}_{\text{constant}} + \underbrace{\frac{1}{2} \mathbf{y}^\top (\mathbf{Q}_{nn} + \sigma_{\text{noise}}^2 I)^{-1} \mathbf{y}}_{\text{data fit}} \\
    &+ \underbrace{\frac{1}{2} \log |\mathbf{Q}_{nn} + \sigma_{\text{noise}}^2 I|}_{\text{complexity}} - \underbrace{\frac{1}{2\sigma_{\text{noise}}^2} Tr(\mathbf{K}_{nn} - \mathbf{Q}_{nn})}_{\text{trace}}\,, \\
\end{aligned}
\end{equation}
where $\mathbf{Q}_{nn} = \mathbf{K}_{nm} \mathbf{K}_{mm}^{-1} \mathbf{K}_{mn}$ is the Nystr\"{o}m approximation to $\mathbf{K}_{nn}$. The separate terms have the following interpretations. The data fit term penalizes inaccuracies in outputs. The complexity term penalizes local density in inducing point selection. When the trace term approaches zero, the $m$ inducing points become a sufficient statistic for the $n$ training samples, meaning that SGP with inducing points $\mathbf{X}_m$ has the same distribution on the   $\mathbf{X}_n$ samples points. 
\section{Method}
\label{sec:method}
Our approach replaces the support neighborhoods of Virani et al.~\cite{ViraniIY20} with a SGP; thus, reducing the computational cost and memory footprint from $n$ data points to $m$ inducing points. This greatly reduces the required computational resources when $m \ll n$, with $m$ tunable based upon validation performance.   

A review of the concepts in Virani et al.~\cite{ViraniIY20} motivates our SGP approach. They first calculate the activations of each of the layers for the training data points $\mathbf{X}_n$. Given a new data point, if there are enough training point activations within an $\epsilon$-neighborhood around the new data point, then there is local evidence to support the final classification decision. If the number of points within the ball is defined as a decision supporting set, then there is support-set-sufficiency to trust the output. At wrost, though, this requires a comparison of all new operating points to all training points.

Instead of using $\epsilon$-neighborhoods to calculate sufficient evidence, our approach includes a probability based measure for sufficient evidence. Specifically, if the training set activation density is estimated using a Gaussian process, then the Gaussian process's training index set and covariance provides an estimate of the local supporting evidence. However, the computations required for a Gaussian process requires a large memory and computational load. We overcome this computational burden using SGPs.     

The SGP's inducing points $\mathbf{X}_m$ are a set of locations that can be used to effectively approximates the Gaussian distribution of the training data set $\mathcal{D}$. The SGP calculates the covariance using only these inducing points and a test sample exploits this computational efficient covariance to predict the support-set-sufficiency. 

\begin{figure}[ht!]
    \centering
    \includegraphics[width=\linewidth]{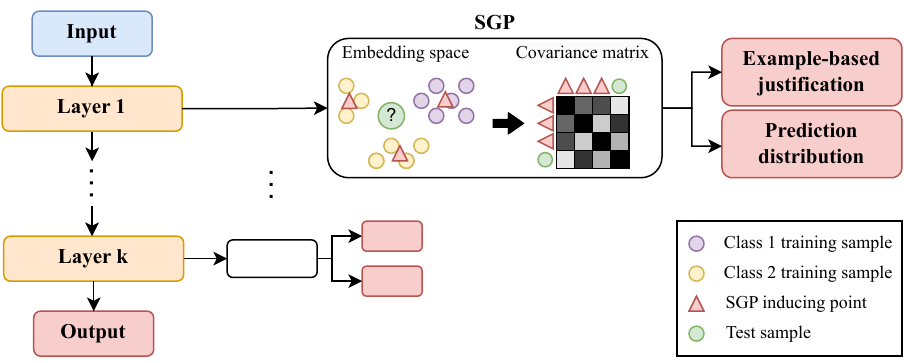}
    \vspace{0.2cm}
    \caption{An illustration of the proposed example-based XAI approach.}
    \label{fig:sgp_epi_classifiers}
\end{figure}

The covariances determines the limits of the support neighborhood; it gives example-based justification for the model’s predictions using the inducing points. Compared with GPs, SGPs reduce the computation cost from $\mathcal{O}(n^3)$ to $\mathcal{O}(nm^2+m^3)$; Thus, our support neighborhood determination is significantly faster than the prior approach for practical use. Figure~\ref{fig:sgp_epi_classifiers} illustrates our approach.

To define the specific calculations, consider $r$ observations $\mathbf{X}_r$, define $a = \min(\mathbf{K}_{mm})$, $b = \max(\mathbf{K}_{mm})$, and $\mathbf{X} = \text{concat}[\mathbf{X}_r, \mathbf{X}_m]$.  We estimate the distance to each inducing point, the covariance and posterior likelihood of generation corresponding to the embedding around that inducing point, and a covariance-adjusted distance: 
\begin{equation}
\begin{aligned}
    \mathbf{D}_{rm} &= \text{pairwise-distance}(\mathbf{X}_r, \mathbf{X}_m), \\
    \mathbf{K} &= k_\mathbf{y}^q(\mathbf{X}, \mathbf{X}), \\
    \mathbf{K}_{rm} &= \frac{\mathbf{K}^{-1}[:r, r:] - a}{b-a}, \\
    \mathbf{P}_{rm} &= \text{clip}(\mathbf{K}_{rm}, 0, 1) \\
    \mathbf{D}_{rm}^\text{cov} &= \mathbf{D}_{rm} + \lambda \mathbf{P}_{rm} \\
\end{aligned}
\label{eqn:metric}
\end{equation}
With $\epsilon$ as a threshold parameter, for each new sample in $\mathbf{X}_r$ counting the number of inducing points such that $\mathbf{D}_{rm} < \epsilon$ or $\mathbf{D}_{rm}^\text{cov} < \epsilon$ provide a means to define $\epsilon$-neighborhoods and support sets. These neighborhoods give local evidence to support the final classification decision. As in Virani et al., this procedure defines epistemic operating points based label-coherent support from the estimated local evidence. At a threshold level, epistemic uncertainty becomes apparent in the presence of label-conflicting inclusions. Moreover, assuming a sufficient number of samples, for inducing points with conflicting labels the function $\mathbf{P}_{rm}$ directly estimate label prediction uncertainty. We showcase these uses below.
\section{Experiments}

We compare the efficacy of our approach to Virani et al.'s performance. Using the CIFAR-10 dataset~\cite{cifar10}, we applied Virani et al.'s method that uses all $n$ data points against our method that uses SGPs and $m<<n$ inducing points using three metrics: final labeling accuracy; new label confidence; and computation time. The new label confidence requires explanation. 

\begin{figure}[hptb!]
    \centering
    \vspace{-5cm}
    \includegraphics[width=0.9\linewidth]{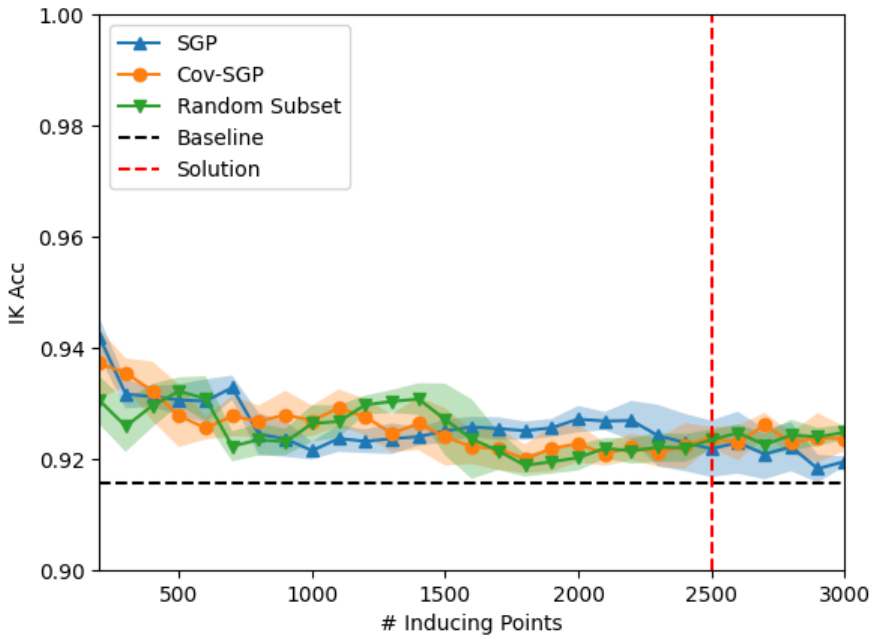}
    \vspace{-5cm}
    \caption{Label accuracy for CIFAR-10}
    \label{acc}
\end{figure}

\begin{figure}[hptb!]
    \centering
    \vspace{-5cm}
    \includegraphics[width=0.9\linewidth]{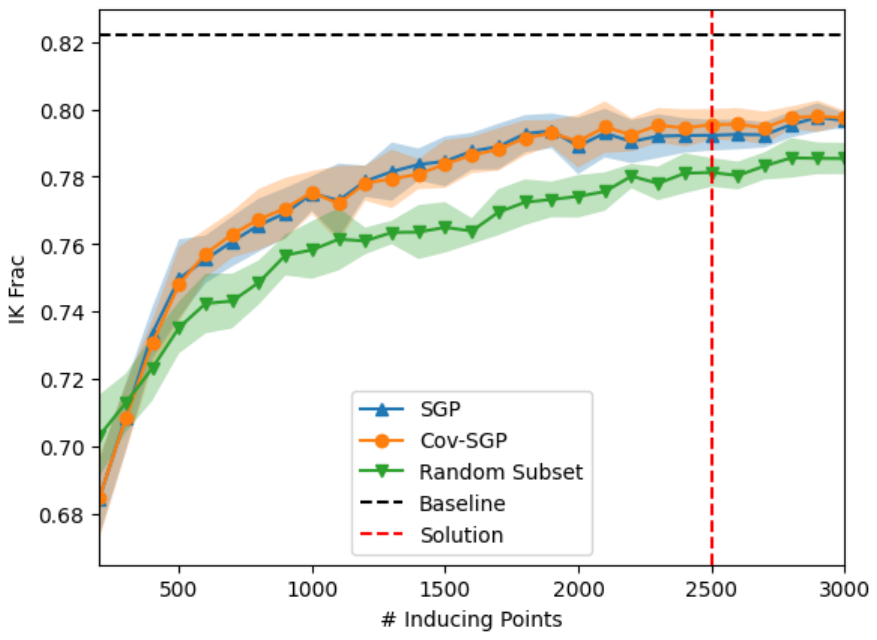}
    \vspace{-5cm}
    \caption{Epistemic operation for CIFAR-10}
    \label{perc}
\end{figure}

\begin{figure}[hptb!]
    \centering
    \vspace{-5cm}
    \includegraphics[width=0.9\linewidth]{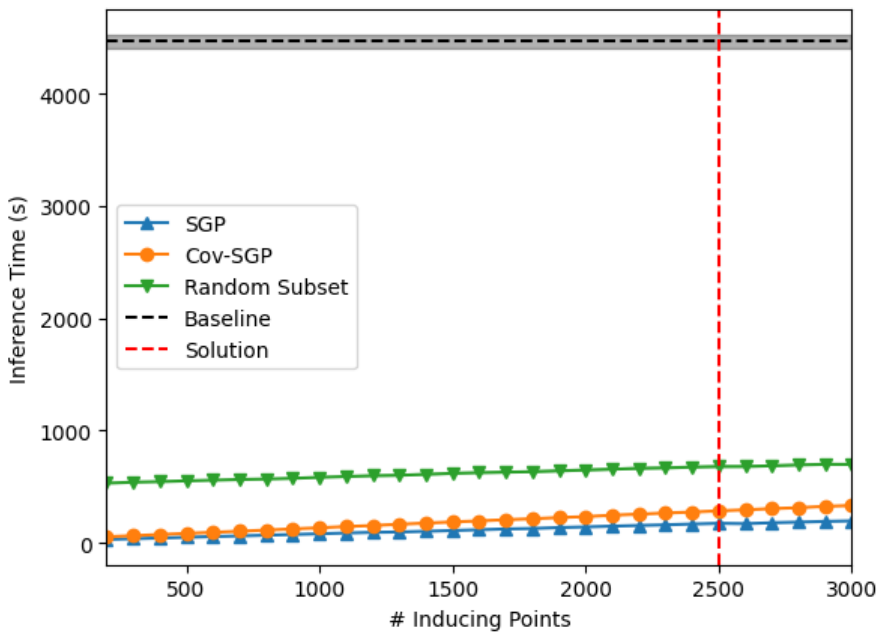}
    \vspace{-5cm}
    \caption{Inference time for CIFAR-10}
    \label{exec}
\end{figure}

New label confidence measures the supporting evidence of a new data point's label and is measured using tunable\\$\epsilon$-neighborhoods. For Virani et al., the $\epsilon$-neighborhood was an $\epsilon$-ball around the new points. For our case, the $\epsilon$-neighborhood is defined by counting the number of inducing points with $D^{\text{cov}}_{rm}$ or $D_{rm}$ in Equation~\ref{eqn:metric}. For both methods, if there are significant (thresholded) number of label-coherent points within the respectively defined $\epsilon$-neighborhoods, then there is sufficient training samples to support the decision and the sample point passes the ''I Know'' (IK) requirement.     

The metrics are shown in three figures. The accuracy of sample points that pass the IK requirement is in Figure \ref{acc}, the fraction of points that pass the IK requirement is in figure \ref{perc}, and execution time is shown in Figure \ref{exec}. In each case, performance compared with the Virani et al. ''Baseline'' is shown via a black dashed line. The ''SGP'' method uses $\mathbf{D}_{rm}$ and ''Cov-SGP'' uses $\mathbf{D}_{rm}^\text{cov}$, and, instead of using variational methods to find the inducing points, ''Random Subset'' uses a randomly selected set of inducing points. The addition of a random subset highlights the utility of including the inducing points as variational parameters in the Sparse Gaussian Process versus less nuanced embeddings. Performance variance over 10 random restarts is shown as a shaded region around each operating curve to help clarify confidence in the application of the approach even with varied instantiations. We observed that there was no significant measurable benefit if the number of inducing points was incrased past 2500. For ease of visual comparison, this plateau effect of epistemic inclusion thresholding is indicated with a dashed red line.

As shown, our SGP or Cov-SGP solution provides assured label performance slightly improving upon the baseline regardless of the number of inducing points selected. Using the IK requirements as an epistemic selection criteria restricts the CIFAR-10 dataset to label 79\% of all validation points vs. 82\% for Virani et al. Of most importance, however, is the reduction of time required for inference by two orders of magnitude; this goes hand-in-hand with the reduction of data points which need to be retained from the original 50,000 by at least an order of magnitude.
\section{Conclusions}

Based upon our findings, we believe that there is strong potential in deploying epistemic uncertainty-aware systems for safe practice even in extremely confined size, weight, and power environments. The efficacy of SGPs for use in embedding training experience suggests an initial path to high-confidence use, while we intend to further explore alternate embeddings and other improvements in future work.

\bibliography{aaai23} % bibliography data in report.bib

\begin{thebibliography}{10}

\bibitem{ViraniIY20}
Virani, N., Iyer, N., and Yang, Z., ``Justification-based reliability in
  machine learning,'' in [{\em Proceedings of the AAAI Conference on Artificial
  Intelligence}{\nolinebreak\hspace{0.1em}]},   {\bf 34}(04),  6078--6085
  (2020).

\bibitem{SnelsonG06}
Snelson, E. and Ghahramani, Z., ``{Sparse Gaussian Processes using
  Pseudo-inputs},'' in [{\em Advances in Neural Information Processing
  Systems}{\nolinebreak\hspace{0.1em}]},  Weiss, Y., Sch\"{o}lkopf, B., and
  Platt, J., eds.,  {\bf 18}, MIT Press (2006).

\bibitem{BuiYT17}
Bui, T.~D., Yan, J., and Turner, R.~E., ``{A Unifying Framework for Gaussian
  Process Pseudo-Point Approximations Using Power Expectation Propagation},''
  {\em Journal of Machine Learning Research}~{\bf 18}(104),  1--72 (2017).

\bibitem{Titsias09}
Titsias, M., ``{Variational Learning of Inducing Variables in Sparse Gaussian
  Processes},'' in [{\em Proceedings of the Twelth International Conference on
  Artificial Intelligence and Statistics}{\nolinebreak\hspace{0.1em}]},  van
  Dyk, D. and Welling, M., eds.,  567--574, PMLR, Florida, USA (2009).

\bibitem{HoangHL15}
Hoang, T.~N., Hoang, Q.~M., and Low, B. K.~H., ``{A Unifying Framework of
  Anytime Sparse Gaussian Process Regression Models with Stochastic Variational
  Inference for Big Data},'' in [{\em Proceedings of the 32nd International
  Conference on Machine Learning}{\nolinebreak\hspace{0.1em}]},  Bach, F. and
  Blei, D., eds.,  {\bf 37},  569--578, PMLR, Lille, France (2015).

\bibitem{SchwalbeF23}
Schwalbe, G. and Finzel, B., ``{A comprehensive taxonomy for explainable
  artificial intelligence: a systematic survey of surveys on methods and
  concepts},'' {\em Data Mining and Knowledge Discovery}  (2023).

\bibitem{AnconaCOG19}
Ancona, M., Ceolini, E., {\"O}ztireli, C., and Gross, M.,  [{\em Gradient-Based
  Attribution Methods}{\nolinebreak\hspace{0.1em}]},  169--191, Springer
  International Publishing, Cham (2019).

\bibitem{Neely2021}
Neely, M., Schouten, S.~F., Bleeker, M. J.~R., and Lucic, A., ``Order in the
  court: Explainable {AI} methods prone to disagreement,'' {\em CoRR}~{\bf
  abs/2105.03287} (2021).

\bibitem{suarez2021}
Suarez, J.~L., Garcia, S., and Herrera, F., ``Ordinal regression with
  explainable distance metric,'' {\em Machine Learning}~{\bf 110},  2729--2762
  (2021).

\bibitem{CandelaRW07}
Quinonero-Candela, J., Rasmussen, C.~E., and Williams, C. K.~I.,
  ``{A}pproximation {M}ethods for {G}aussian {P}rocess {R}egression,'' in [{\em
  Large-Scale Kernel Machines}{\nolinebreak\hspace{0.1em}]},   203--223, MIT
  Press (2007).

\bibitem{WilkDJAAH20}
{van der Wilk}, M., Dutordoir, V., John, S., Artemev, A., Adam, V., and
  Hensman, J., ``{A Framework for Interdomain and Multioutput Gaussian
  Processes},'' {\em ArXiv}  (2020).

\bibitem{BorovitskiyTMD20}
Borovitskiy, V., Terenin, A., Mostowsky, P., and Deisenroth~(he/him), M.,
  ``Mat\'{e}rn {G}aussian processes on riemannian manifolds,'' in [{\em
  Advances in Neural Information Processing
  Systems}{\nolinebreak\hspace{0.1em}]},  Larochelle, H., Ranzato, M., Hadsell,
  R., Balcan, M., and Lin, H., eds.,  {\bf 33},  12426--12437, Curran
  Associates, Inc. (2020).

\bibitem{RasmussenW05}
Rasmussen, C.~E. and Williams, C. K.~I.,  [{\em {Gaussian Processes for Machine
  Learning}}{\nolinebreak\hspace{0.1em}]}, MIT Press, Cambridge, USA (2005).

\bibitem{BurtRW19}
Burt, D., Rasmussen, C.~E., and Van Der~Wilk, M., ``{Rates of Convergence for
  Sparse Variational {G}aussian Process Regression},'' in [{\em Proceedings of
  the 36th International Conference on Machine
  Learning}{\nolinebreak\hspace{0.1em}]},  Chaudhuri, K. and Salakhutdinov, R.,
  eds.,  {\bf 97},  862--871, PMLR (Jun 2019).

\bibitem{BauerWR16}
Bauer, M., van~der Wilk, M., and Rasmussen, C.~E., ``{Understanding
  Probabilistic Sparse Gaussian Process Approximations},'' in [{\em Advances in
  Neural Information Processing Systems}{\nolinebreak\hspace{0.1em}]},
  1533–1541 (2016).

\bibitem{cifar10}
Krizhevsky, A., ``Learning multiple layers of features from tiny images,''
  (2009).

\end{thebibliography}
\bibliographystyle{spiebib} % makes bibtex use spiebib.bst

\end{document}